\documentclass[utf8]{FrontiersinVancouver} 

\usepackage{url,hyperref,microtype,subcaption}
\usepackage[onehalfspacing]{setspace}
\makeatletter
\patchcmd{\ps@headings}{\hbox to \textwidth{\helveticabold\small {Frontiers}\hfill \thepage}}{\hbox to \textwidth{\hfill \helveticabold\small \thepage}}{}{}
\makeatother
\usepackage{booktabs}
\usepackage{longtable}
\usepackage{array}
\usepackage{ragged2e}
\usepackage{multirow}
\usepackage{graphicx}

\makeatletter
\patchcmd{\@maketitle}{\flushleft \includegraphics[width=12pc,angle=0]{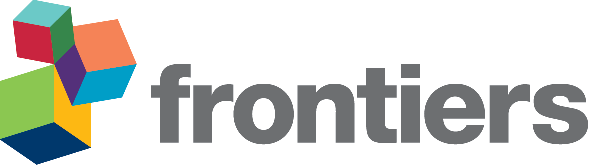}}{}{}{}
\patchcmd{\ps@headings}{\helveticabold\small {Frontiers}\hfill \thepage}{\helveticabold\small \hfill \thepage}{}{}
\patchcmd{\ps@headings}{\helveticabold\small {Frontiers}\hfill \thepage}{\helveticabold\small \hfill \thepage}{}{}
\makeatother
\newcolumntype{L}[1]{>{\RaggedRight\arraybackslash}p{#1}}
\newcolumntype{R}[1]{>{\RaggedLeft\arraybackslash}p{#1}}

\def\keyFont{\fontsize{8}{11}\helveticabold }
\def\firstAuthorLast{Hartsock {et~al.}} 
\def\Authors{Iryna Hartsock\,$^{1,*}$, Cesar Lam\,$^{2}$, Christopher Otteni\,$^{2}$, Aliya Qayyum\,$^{2}$, Robert Gatenby\,$^{2}$, Cyrillo Araujo\,$^{2}$, and Ghulam Rasool\,$^{1,2, 3, 4}$}
\def\Address{$^{1}$ Department of Machine Learning, H. Lee Moffitt Cancer Center \& Research Institute, Tampa, FL, USA \\
$^{2}$ Department of Diagnostic Imaging and Interventional Radiology, H. Lee Moffitt Cancer Center \& Research Institute, Tampa, FL, USA \\
$^{3}$ Department of Neuro-Oncology, H. Lee Moffitt Cancer Center \& Research Institute, Tampa, FL, USA \\
$^{4}$ Department of Translational Pathology, H. Lee Moffitt Cancer Center \& Research Institute, Tampa, FL, USA }
\def\corrAuthor{Iryna Hartsock}

\def\corrEmail{iryna.hartsock@moffitt.org}

\begin{document}
\onecolumn
\firstpage{1}

\title[Multi-Agent AI System for Radiology Report Structuring and Quality Assurance]{Multi-Agent AI System for Radiology Report Structuring and Quality Assurance with Independent Radiologist Evaluation} 

\author[\firstAuthorLast ]{\Authors} 
\address{} 
\correspondance{} 

\extraAuth{}

\maketitle

\begin{abstract}

Purpose: To develop and evaluate a locally deployed multi-agent AI system for radiology report structuring and quality assurance.
Materials and Methods: This retrospective study included 638 radiology reports from CT examinations of the chest, abdomen, and pelvis dictated by 15 board-certified radiologists in 2023 and 2024. A multi-agent AI pipeline was developed to perform report structuring and quality assurance (QA).  The system structured the report into standardized anatomical sections at the sentence level using regex rules and local large language models. It also detected mismatches between the Findings and Impression sections, or within sections; gender–anatomy conflicts; and undocumented communication of critical findings. Two board-certified radiologists independently evaluated a 45-report subset.
Results: The multi-agent system structured the Findings sections of all reports (22,270 sentences) into a predefined anatomical format while retaining the original report content. The system flagged 90 (14.1\%) reports, most commonly for section mismatches (80 reports, 12.5\%). In the radiologist evaluation, both reviewers agreed that 31 (69\%) were correctly restructured, 2 reports (4\%) were incorrectly restructured, and disagreed on the remaining 12 reports (27\%). Both reviewers agreed that no clinically important information was omitted and no fabricated content was introduced. Overall QA performance was rated as ``excellent'' or ``good'' in 84\% of the evaluated reports, with the remaining reports rated as ``fair''.
Conclusion: A locally deployed multi-agent AI system combined radiology report structuring and quality assurance within a single workflow. The system demonstrated favorable performance in radiologist evaluation. Such systems may support standardization of reporting and quality assurance in radiology practice.

\tiny
 \keyFont{ \section{Keywords:} multi-agent artificial intelligence, radiology reports, structured reporting, quality assurance, large language models} 
\end{abstract}

\section{Introduction}
A challenge in radiology reporting is inconsistent or absent structure, which can lead referring physicians to overlook critical clinical information or spend time searching for it \cite{Weiss2008StructuredReporting}. Radiology reports include sections such as findings and impressions, with findings sometimes described by anatomic regions or organs.  However, the organization and placement of findings within reports can vary across radiologists, which may complicate communication between radiologists and referring clinicians \cite{Bosmans2011RadiologyReport, Sistrom2005Framework}. In addition, radiology reports without a consistent structure can limit their secondary use for research, quality monitoring, and clinical decision support systems \cite{Pons2016NLP}. 

Several approaches have been proposed to address these limitations. Structured reporting templates have been developed to standardize radiology reporting and improve consistency, although adoption has been variable due to concerns regarding workflow burden and reporting flexibility \cite{Weiss2008StructuredReporting, Nobel2022review}. Recently, natural language processing (NLP) and large language model (LLM) approaches have been explored to extract structured information from narrative radiology reports or convert free-text reports into structured formats \cite{Adams2023GPT4StructuredReporting, Hartsock2026Conciseness, Woznicki2025structed_reporting}. These methods typically focus on extracting specific findings, generating labels, or summarizing report content and do not preserve the full content of the original report.

Another challenge is the presence of reporting discrepancies or communication errors that can lead to misinterpretation, delayed treatment, or inappropriate clinical management \cite{Brady2017ErrorDiscrepancy}. In some cases, inconsistencies within a report may require clinicians to re-examine imaging studies for clarification. The widespread adoption of voice recognition systems further highlights the need for quality control, as speech-to-text reporting may introduce transcription errors or unintended wording changes \cite{McGurk2008VoiceRecognition}. Automated methods can more efficiently detect such inconsistencies than manual review, particularly in high-volume reporting environments. Recent studies have explored the use of LLMs for automated error detection in radiology reports, demonstrating that LLMs can identify certain types of inconsistencies within report text \cite{Sun2025GenerativeLLMErrors, Gertz2024GPT4Errors, Salam2025LLMErrorDetection, Kim2025GPT4Proofreading, Kim2026error-detection}, representing an emerging and rapidly evolving direction in clinical AI research \cite{Hartsock2026ClinicalAI}. 

In this work, we propose a multi-agent AI system that standardizes radiology report organization and performs automated quality assurance (QA) while preserving the original report text. The system assigns sentences from the report to predefined anatomical sections without rewriting the content. It also identifies inconsistencies, including mismatches between findings and impressions, discrepancies within sections, gender–anatomy conflicts, and critical findings without documented communication. This approach enables structured reporting and automated quality checks while allowing radiologists to maintain their reporting style. A subset of reports was evaluated by radiologists to assess the quality of the multi-agent AI system.

\section{Materials and Methods}

\subsection{Radiology Reports}

In this retrospective study, we collected 638 complete radiology reports based on CT scans of the chest, abdomen, and pelvis. Reports were dictated by 15 board-certified radiologists using a voice-recognition system between 2023 and 2024. The reports ranged from 146 to 1,594 words, with an average of 438 words. Fourteen radiologists organized the Findings section by organ-based sections using different styles, while reports from one radiologist had largely unstructured Findings sections. Radiology reports were processed behind the institution's firewall using locally deployed AI models. The institutional review board (IRB) approved the study and waived the requirement for informed consent from participating radiologists and patients.

\subsection{Report Preprocessing for the Structuring Task}
For the structuring task, the Findings section was extracted using rule-based pattern matching to identify section boundaries. A custom sentence segmentation algorithm was implemented to handle radiology-specific formatting, including numbered lists, multi-line sentences, and punctuation inconsistencies. Short follow-up fragments (e.g., ``Stable since prior'') were conditionally merged with the preceding sentence using rule-based pattern matching to preserve contextual meaning. These steps produced sentence-level inputs for downstream organ-based structuring of the report.  

\begin{figure}[ht]
\begin{center}
\includegraphics[width=17cm]{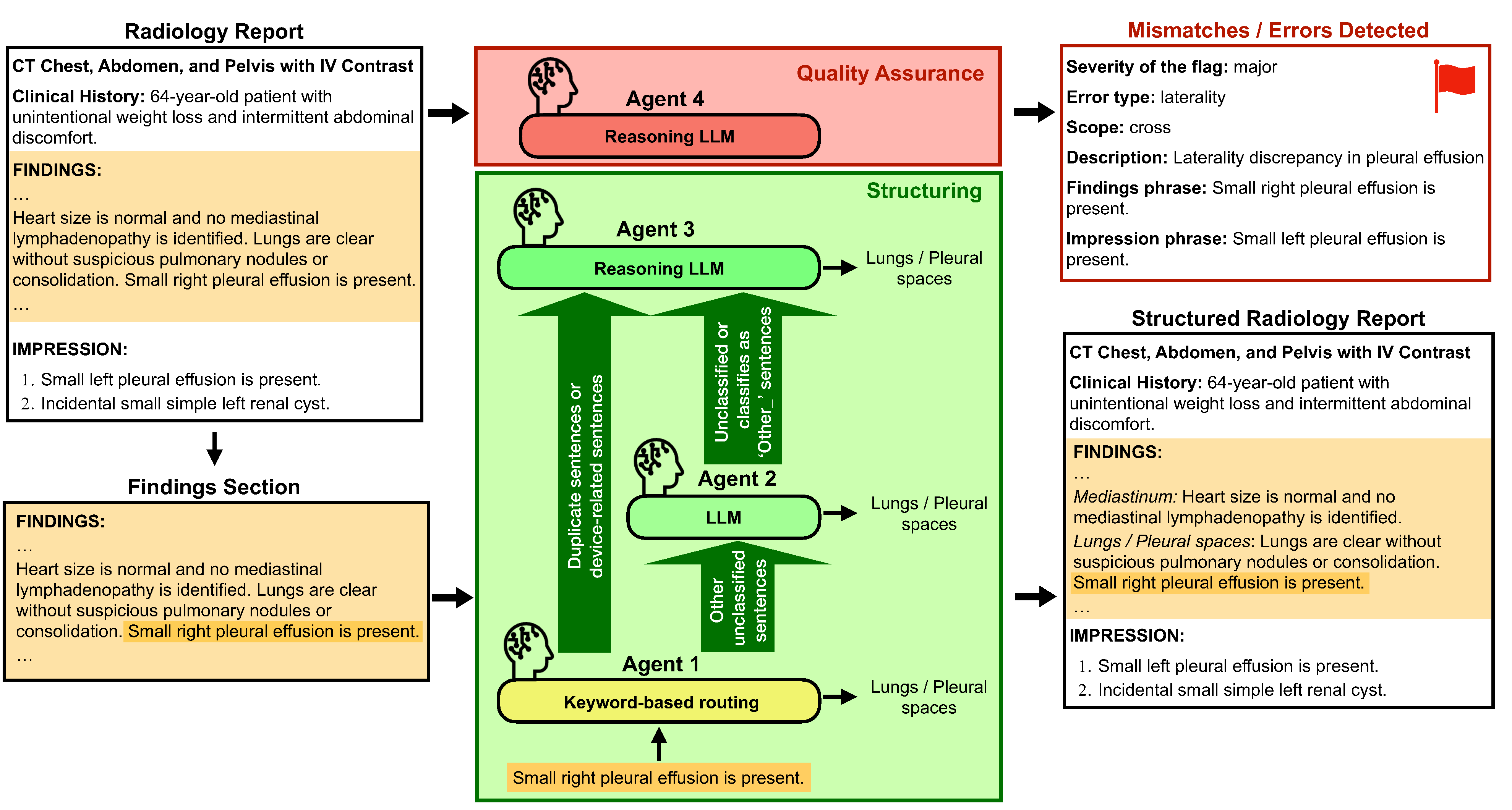}
\end{center}
\caption{ Overview of the proposed multi-agent AI pipeline for radiology report structuring and quality assurance. The structuring task operates at the sentence level and reorganizes Findings section content into predefined anatomical sections. Each sentence is first evaluated by the rule-based agent (Agent 1). Device-related sentences and duplicate sentences not classified by Agent 1 are routed directly to the DeepSeek-R1 reasoning model (Agent 3). Other unclassified sentences are processed by the LLaMA-3 model (Agent 2), and sentences that remain unclassified or are assigned to nonspecific ``Other'' categories are subsequently reviewed by Agent 3. Quality assurance is performed at the whole-report level by a separate DeepSeek-R1 agent (Agent 4), which flags potential reporting inconsistencies and assigns a severity level to the flag. The upper-right panel shows an example output for a Findings–Impression mismatch; similar outputs are generated for gender–anatomy discrepancies and undocumented communication of critical findings.}\label{fig:pipeline}
\end{figure}

\subsection{Multi-agent AI system pipeline}
\begin{figure}[ht]
\begin{center}
\includegraphics[width=18cm]{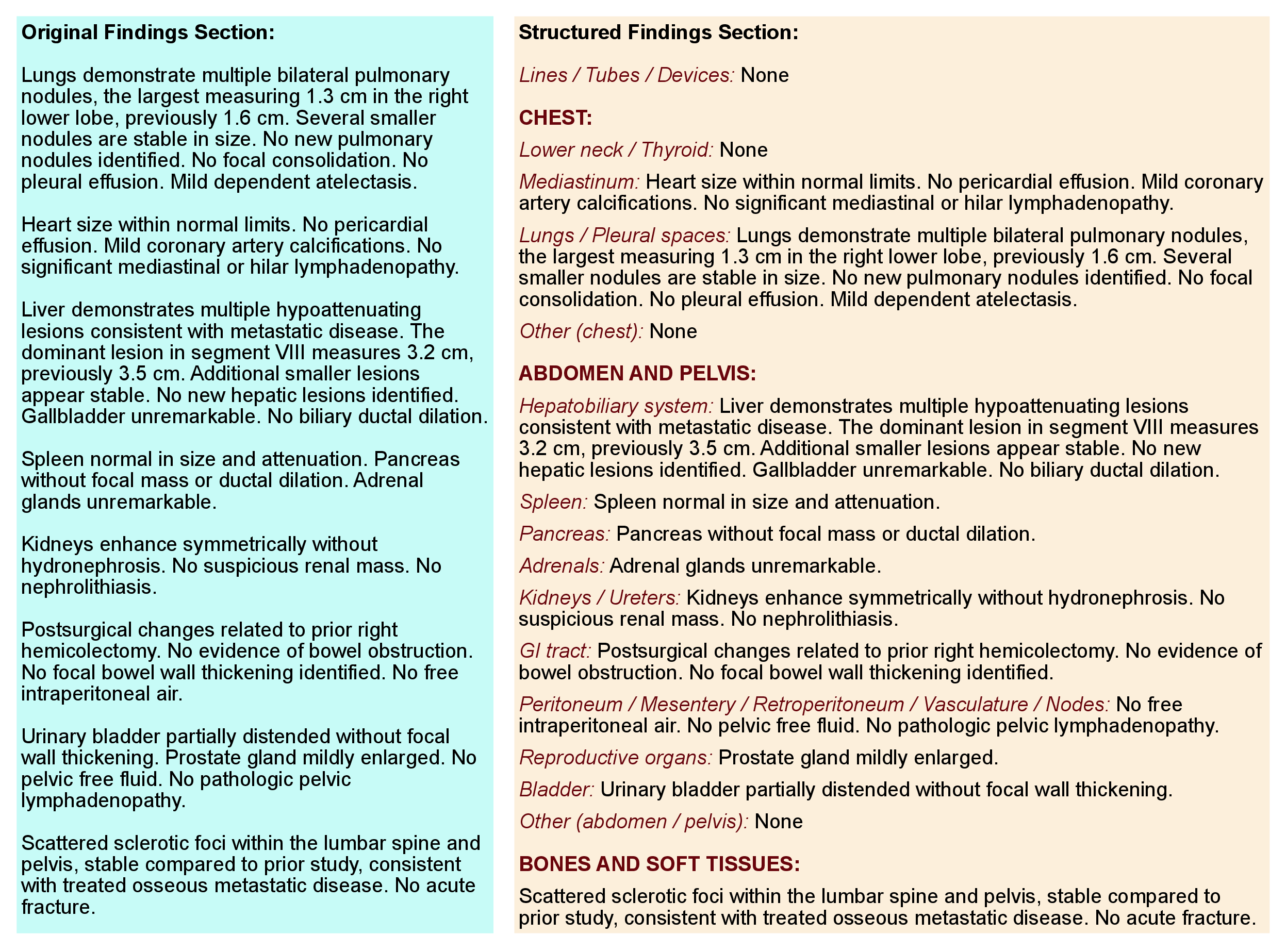}
\end{center}
\caption{Example of automated structuring of the radiology report Findings section by the proposed multi-agent AI system. The left panel shows the original Findings section from a synthetic radiology report. The right panel shows the system's corresponding structured output, with findings reorganized into pre-defined organ-based sections. All Findings sections of the radiology reports processed by the pipeline were converted to this structured format.}\label{fig:example}
\end{figure}

We developed a four-agent pipeline for radiology report structuring and QA, shown in Figure \ref{fig:pipeline}. The first three agents organized findings into the predefined anatomical format shown in Figure \ref{fig:example}, while the fourth agent performed report-level QA. The first agent performed deterministic sentence classification using a domain-specific dictionary of approximately 600 regular expression (regex) patterns covering common anatomical regions and radiological findings. The classification rules and anatomical assignments were developed with input from a board-certified radiologist to classify common findings into predefined anatomical sections. The second agent applied an LLM (LLaMA-3–8B, quantized to q8) \cite{Grattafiori2024Llama3} to resolve cases not captured by rules, using the full Findings text and local sentence context to infer organ-level associations. The third agent employed a reasoning LLM (DeepSeek-R1-Distill-Llama-70B, quantized to q4) \cite{Guo2025DeepSeekR1} for selected sentences requiring additional contextual interpretation. Device-related sentences were identified using device-specific regex patterns and routed directly to DeepSeek-R1 because they often required assignment to both the “Devices” section and the relevant anatomical section describing the device location or associated finding. Duplicate sentences that Agent 1 could not classify were also bypassed by the LLaMA agent and routed directly to DeepSeek-R1. In addition, DeepSeek-R1 reviewed sentences that remained unclassified or were assigned to “Other” categories by the LLaMA agent. In all cases, the sentence being classified was provided together with the full Findings section.

The fourth agent, also based on DeepSeek-R1-Distill-Llama-70B (quantized to q4), operated at the whole-report level to identify clinically meaningful inconsistencies, including cross-section and within-section laterality discrepancies, negation reversals, polarity conflicts, organ or segment mismatches, size or unit discrepancies, gender–anatomy inconsistencies, and undocumented communication of potentially critical findings, while suppressing clinically acceptable summarization differences between Findings and Impressions. The QA agent also assigned an overall severity level (none, minor, major, or critical) to each report. 

All LLM-based agents used structured prompts tailored to their respective tasks and generated predefined JSON outputs for automated downstream processing. Structuring agents performed single-pass inference without retry loops, whereas the QA agent used a retry mechanism of up to five attempts when strict JSON parsing failed. The pipeline was implemented in Python and executed on a computing cluster using a single NVIDIA H100 GPU and 12 CPU cores. Local LLM inference was performed through the Ollama \footnote{\url{https://ollama.com/} (accessed June 4, 2026)} platform, which served the quantized LLaMA-3-8B and DeepSeek-R1-Distill-Llama-70B models. The multi-agent pipeline is available at: \href{https://github.com/lab-rasool/Multi-agent-AI-system-for-Radiology-Report-Structuring-and-Quality-Assurance}{https://github.com/lab-rasool/Multi-agent-AI-system-for-Radiology-Report-Structuring-and-Quality-Assurance}.

\subsection{Expert Evaluation Process}
Clinical evaluation was conducted through independent review by 2 board-certified radiologists using a standardized evaluation form. The evaluating radiologists were distinct from the group that authored the original reports. A fixed set of 45 reports was sampled and provided to all reviewers, consisting of three reports per radiologist-writer (with author identities redacted). For each radiologist-writer, up to three reports flagged by the agentic system were randomly selected; if fewer than three were available, all flagged reports were included, and the remaining cases were randomly sampled from unflagged reports to maintain three reports per radiologist-writer. This sampling procedure yielded 41 flagged and 4 unflagged reports. The review focused on both the agentic system's structuring and QA components. Reviewers assessed whether sentence-level classifications were correctly assigned to the appropriate organ-based sections, whether any clinically relevant information was omitted, or any fabricated content was introduced during structuring, and whether the structured report was preferable to the original report for clinical review. Reviewers also evaluated whether system-flagged errors were correct and complete, whether severity assignments were appropriate, whether the reasoning generated by the QA LLM agent was logically consistent and clinically sound, and the overall performance of the QA agent. Unless otherwise specified, reported evaluation outcomes in the Results section represent agreement between both reviewers.

\section{Results}
\subsection{Radiology Report Structuring}
Sentence-level classification of Findings section content across 638 radiology reports (22,270 sentences) processed by the multi-agent AI system is summarized in Table \ref{tab:sentence_classification}. Most sentences were handled by the rule-based stage, with Agent 1 (regex) classifying 18,493 sentences (83.0\%). Agent 2 (LLaMA 3) accounted for 2,729 sentences (12.3\%), primarily capturing cases where simple pattern matching was insufficient, while Agent 3 (DeepSeek-R1) classified 1,041 sentences (4.7\%) that required additional reasoning or disambiguation. Only 7 sentences ( $<$ 0.04\%) remained unclassified; these were not true misses but report section headers. Because reports did not consistently use punctuation, sentence boundaries were determined using a combination of delimiters (e.g., line breaks and spacing), which occasionally isolated headers as standalone sentences.

Performance was generally consistent across radiologists, with some variability in how often later agents were required. Reports written by the radiologists 3, 6, and 13 had the highest proportion of sentences classified by Agent 1 (93.3\%, 90.0\%, and 90.3\%, respectively), indicating strong alignment with the rule-based patterns. In contrast, reports from radiologists 2 and 4 showed the greatest usage of LLM-based agents (26.7\% and 27.2\% combined for Agents 2 and 3), reflecting increased reliance on contextual classification. Reports from radiologist 11 showed the highest use of Agent 3 (12.7\%), suggesting a greater proportion of sentences requiring deeper reasoning or disambiguation compared with other radiologists.

\subsection{Radiology Report Quality Assurance}
The counts of various QA flags across 638 reports are shown in Table \ref{tab:qa_flags}. A total of 90 reports (14.1\%) were flagged by Agent 4 (DeepSeek-R1) for potential inconsistencies or errors. Most flags were non-critical: 40 reports (6.3\%) contained major severity findings, 44 (6.9\%) contained minor findings, and only 6 (0.9\%) were classified as critical. The most common error type was section mismatches, identified in 80 reports (12.5\%), indicating discrepancies between Findings and Impression sections or within either section. Gender-anatomy discrepancies were rare (4 reports, 0.6\%), as were missing critical findings communication flags (9 reports, 1.4\%). Across radiologists, the proportion of flagged reports varied, ranging from 4.8\% (radiologist 3) to 27.5\% (radiologist 1). Radiologists 1, 6, and 9 showed higher flagging rates (27.5\%, 22.2\%, and 24.1\%, respectively), while others such as radiologists 3 and 14 had relatively low rates (4.8\% and 5.0\%).

\subsection{Processing Time}
The average processing time per report for each stage of the pipeline is shown in Table \ref{tab:processing_time}. The overall average total processing time was 55.6 seconds per report. The regex-based stage (Agent 1) was the fastest component (0.54 seconds per report), followed by Agent 2 (1.31 seconds). In contrast, the majority of processing time was attributed to the DeepSeek-R1 agents, with Agent 3 averaging 28.62 seconds and Agent 4 averaging 25.13 seconds per report. Processing time varied across radiologists, with the longest average runtime observed for radiologist 15 (265.5 seconds). This was due to increased Agent 3 processing time, as radiologist 15's reports contained poorly organized findings by anatomical sections, requiring more contextual interpretation by the reasoning model.

\subsection{Radiologist Evaluation of the Performance of the Multi-Agent AI System}

Independent evaluations performed by two radiologists across 45 (7\% of all reports) reports for both the structuring and QA tasks are summarized in Table \ref{tab:radiologist_evaluation}. For report restructuring, both reviewers agreed that 31 reports (69\%) were correctly restructured, while 2 reports (4\%) were incorrectly restructured, meaning at least 1 sentence was not correctly classified. Reviewer disagreement occurred in 12 reports (27\%), primarily involving borderline section assignment or organizational preferences. Both reviewers agreed that no clinically important information was omitted and no fabricated information was introduced in any evaluated report. For clinical usability, both reviewers rated the restructured report as “about the same” as the original report in 23 cases (51\%), “better” in 1 case (2\%), and “worse” in 3 cases (7\%), with disagreement in 18 cases (40\%), mostly between better and “about the same”. The only report that both radiologists agreed was improved by restructuring was the report whose Findings section was not organized by organ system in the original report.

For QA-related tasks, mismatches between the Findings and Impression sections, or within either section, were correctly identified in 27 reports, while the absence of a mismatch was correctly identified in 10 reports; overall, 37 reports (82\%) had correctly assessed mismatches. False-positive mismatch flags occurred in 5 reports, with evaluator disagreement regarding mismatch assessment in an additional 3 reports. Gender–anatomy discrepancy detection produced 1 incorrect discrepancy flag across all evaluated reports, caused by misidentification of the research coordinator’s name in the indication section as the patient’s name, resulting in an apparent gender mismatch. Detection of undocumented communication of potentially critical findings correctly identified missing communication in 5 reports and correctly identified no communication issue in 38 reports, resulting in correct assessment in 43 reports (96\%). One false-positive flag and one evaluator disagreement were also identified. Severity assignment was considered correct in 28 reports (22 minor and 6 major) and incorrect in 2 reports, with evaluator disagreement in 15 reports. Overall QA performance was rated as ``excellent'' in 21 reports, ``good'' in 2 reports, and ``fair'' in 7 reports, with disagreement between ``excellent'' and ``good'' assessments in 15 reports.

\section{Discussion} 
In this study, we developed and evaluated a multi-agent AI system for radiology report structuring and quality assurance, combining rule-based methods with locally deployed LLMs. The system organized radiology reports written in varying styles into a standardized anatomical structure, preserving the original report text while simultaneously performing automated QA analysis for clinically relevant inconsistencies. Independent radiologist evaluation of the 45-report subset demonstrated favorable performance for both structuring and QA tasks.

By combining deterministic rule-based classification with selective LLM-based reasoning for more ambiguous findings, the pipeline achieved consistent report organization while limiting reliance on computationally intensive models. Previous studies have demonstrated that LLMs can transform free-text radiology reports into structured formats, including with locally deployed open-source models \cite{Adams2023GPT4StructuredReporting, Hartsock2026Conciseness, Woznicki2025structed_reporting}. In contrast to approaches that generate or rewrite structured report content, the present system assigned existing report sentences to anatomical sections without modifying the source text, thereby limiting opportunities for unsupported content generation. Approximately 95\% of sentences were classified by the regex-based and LLaMA-based agents without requiring escalation to the DeepSeek-R1 reasoning model. Both reviewers agreed that 69\% of reports were correctly restructured, while only 4\% were considered incorrectly restructured. Disagreement occurred in 27\% of reports, most often involving findings that could reasonably be assigned to more than one anatomical section because of overlapping anatomy or differences in report organization preferences. Importantly, both reviewers agreed that no clinically important findings were omitted and no fabricated findings were introduced, indicating that the original report content was preserved despite occasional differences in preferred organization. Reviewers most commonly rated the restructured reports as comparable to the original reports rather than superior. This was largely attributable to occasional duplication of findings across multiple anatomical sections when a statement was considered relevant to more than one category. Although this approach reduced the risk of omitting information, duplicated statements sometimes reduced the report's perceived readability.

The QA component identified several types of reporting inconsistencies, predominantly mismatches between Findings and Impression sections, with relatively few gender–anatomy discrepancies or cases of potentially critical findings without documented communication. These findings are consistent with recent studies demonstrating the potential of LLMs to identify errors and internal inconsistencies in radiology reports
\cite{Sun2025GenerativeLLMErrors, Gertz2024GPT4Errors, Salam2025LLMErrorDetection, Kim2025GPT4Proofreading, Kim2026error-detection}.
Previous LLM-based radiology QA studies have evaluated errors such as laterality, negation, transcription, omission, and discrepancies between Findings and  Impression sections. The present QA agent extended this scope by assessing inconsistencies both across and within report sections and by evaluating gender–anatomy inconsistencies and undocumented communication of potentially critical findings. To our knowledge, these latter error categories have not been systematically evaluated in prior LLM-based radiology report QA studies.

In radiologist review, all but one identified mismatch were detected by the QA agent. In one report, one reviewer identified an additional discrepancy involving conflicting references to abdominoperineal resection (APR) and low anterior resection (LAR) that was not flagged by the QA agent or identified by the second reviewer, illustrating that subtle inconsistencies may also be overlooked during manual review. False-positive QA flags occurred in 7 of the 45 evaluated reports. Both reviewers rated overall QA performance as either ``excellent'' or ``good'' in 38 of 45 reports (84\%), although they differed between these two favorable categories in 15 reports; the remaining 7 reports (16\%) were rated as ``fair'' by both reviewers. No reports received a ``poor'' rating. Reviewer assessment indicated that the QA agent’s reasoning was clinically reasonable, with disagreements on distinctions between fully correct and mostly correct reasoning rather than fundamentally incorrect conclusions. Despite requiring approximately 25 seconds per report, the QA agent demonstrated good performance across multiple error categories. 

A potential advantage of the proposed system is the integration of report structuring and quality assurance within a single workflow. Inconsistent organization of radiology reports can make it more difficult for referring physicians to quickly identify and interpret clinically relevant findings. By reorganizing findings into a consistent anatomical framework, the system provides a standardized report structure while preserving the original report content. It also achieved consistent structuring across 15 different radiologist writing styles, underscoring its ability to handle the degree of variability commonly encountered in real-world reporting environments. The QA component further enables automated identification of reporting inconsistencies and communication issues that may otherwise require manual review. In addition, all models were deployed locally behind the institution’s firewall.

This study has several limitations. Clinical evaluation was limited to a review of 45 reports by two radiologists. While a larger evaluation cohort would strengthen the findings, a comprehensive radiologist review of multi-agent outputs would require substantial time and effort. The expert evaluation subset was intentionally enriched for QA-flagged reports (41 flagged and 4 unflagged) and therefore was not intended to provide unbiased estimates of overall QA performance across the full dataset. Some aspects of report organization are also inherently subjective, particularly when findings are relevant to multiple anatomical sections and may reasonably be categorized differently by different radiologists. In a small number of cases, the restructuring process duplicated findings across multiple sections, improving information retention but occasionally reducing readability. Similarly, the QA component was designed to prioritize identifying potential inconsistencies and therefore produced a small number of false-positive flags. Future work should focus on validation across additional institutions, imaging modalities, and reporting styles, as well as further refinement of the structuring and QA components.

In conclusion, the proposed system demonstrated that automated report structuring and quality assurance can be integrated within a single workflow. The combination of deterministic and reasoning-based agents enabled consistent organization of report content and the identification of clinically relevant reporting discrepancies across a heterogeneous dataset. Although further validation is needed, these findings suggest that agent-based AI approaches may help improve reporting consistency and support quality assurance processes in radiology practice.

\section*{Conflict of Interest Statement}

The authors declare that the research was conducted in the absence of any commercial or financial relationships that could be construed as a potential conflict of interest.

\section*{Author Contributions}

Conceptualization, I.H., C.L., C.A., and G.R.; software development and formal analysis, I.H.; LLM prompt refinement and keyword selection, I.H. and C.A.; radiology report evaluation, C.L. and C.O.; writing—original draft, I.H.; funding acquisition, A.Q., R.G., and G.R.;writing—review and editing, all authors. 




\section*{Data Availability Statement}
The radiology report data analyzed in this study are not publicly available because they contain protected health information.

\bibliographystyle{Frontiers-Vancouver} 
\bibliography{references}


\newpage
\begin{table}[ht]
\centering

\caption{Number of reports and total number of sentences in the Findings sections for each radiologist, with sentence-level classification distribution across the three-agent report structuring pipeline, including sentences classified by Agent 1 (regex), Agent 2 (LLaMA 3), Agent 3 (DeepSeek-R1), and remaining unclassified sentences.}
\label{tab:sentence_classification}

\renewcommand{\arraystretch}{1.05}
\setlength{\tabcolsep}{4pt}

\resizebox{\textwidth}{!}{%
\begin{tabular}{@{}ccccccc@{}}

\toprule

\multirow[c]{3}{*}{\textbf{Radiologist}} &
\multirow[c]{3}{*}{\textbf{Reports, n}} &
\multirow[c]{3}{*}{\textbf{Sentences, n}} &
\multicolumn{3}{c}{\textbf{Sentences classified by, n (\%)}} &
\multirow[c]{3}{*}{%
\shortstack[c]{\textbf{Unclassified}\\
\textbf{sentences, n (\%)}}} \\

\cmidrule(lr){4-6}

&
&
&
\shortstack[c]{\textbf{Agent 1}\\\textbf{(regex)}} &
\shortstack[c]{\textbf{Agent 2}\\\textbf{(LLaMA 3)}} &
\shortstack[c]{\textbf{Agent 3}\\\textbf{(DeepSeek-R1)}} &
\\

\midrule

\multirow{2}{*}{Radiologist 1} &
\multirow{2}{*}{40} &
\multirow{2}{*}{1,403} &
1,151 & 187 & 63 & 2 \\
&
&
&
(82\%) & (13.3\%) & (4.5\%) & ($<$0.2\%) \\

\multirow{2}{*}{Radiologist 2} &
\multirow{2}{*}{19} &
\multirow{2}{*}{767} &
562 & 141 & 64 & \multirow{2}{*}{0} \\
&
&
&
(73.3\%) & (18.4\%) & (8.3\%) & \\

\multirow{2}{*}{Radiologist 3} &
\multirow{2}{*}{42} &
\multirow{2}{*}{1,382} &
1,289 & 30 & 61 & 2 \\
&
&
&
(93.3\%) & (2.2\%) & (4.4\%) & ($<$0.2\%) \\

\multirow{2}{*}{Radiologist 4} &
\multirow{2}{*}{45} &
\multirow{2}{*}{2,211} &
1,609 & 440 & 162 & \multirow{2}{*}{0} \\
&
&
&
(72.8\%) & (19.9\%) & (7.3\%) & \\

\multirow{2}{*}{Radiologist 5} &
\multirow{2}{*}{62} &
\multirow{2}{*}{2,148} &
1,769 & 344 & 35 & \multirow{2}{*}{0} \\
&
&
&
(82.4\%) & (16\%) & (1.6\%) & \\

\multirow{2}{*}{Radiologist 6} &
\multirow{2}{*}{27} &
\multirow{2}{*}{1,487} &
1,338 & 112 & 37 & \multirow{2}{*}{0} \\
&
&
&
(90\%) & (7.5\%) & (2.5\%) & \\

\multirow{2}{*}{Radiologist 7} &
\multirow{2}{*}{39} &
\multirow{2}{*}{1,021} &
851 & 132 & 37 & 1 \\
&
&
&
(83.3\%) & (12.9\%) & (3.6\%) & ($<$0.2\%) \\

\multirow{2}{*}{Radiologist 8} &
\multirow{2}{*}{21} &
\multirow{2}{*}{1,192} &
984 & 152 & 56 & \multirow{2}{*}{0} \\
&
&
&
(82.6\%) & (12.8\%) & (4.7\%) & \\

\multirow{2}{*}{Radiologist 9} &
\multirow{2}{*}{29} &
\multirow{2}{*}{948} &
782 & 86 & 80 & \multirow{2}{*}{0} \\
&
&
&
(82.5\%) & (9.1\%) & (8.4\%) & \\

\multirow{2}{*}{Radiologist 10} &
\multirow{2}{*}{11} &
\multirow{2}{*}{599} &
489 & 80 & 30 & \multirow{2}{*}{0} \\
&
&
&
(81.6\%) & (13.4\%) & (5\%) & \\

\multirow{2}{*}{Radiologist 11} &
\multirow{2}{*}{36} &
\multirow{2}{*}{1,120} &
861 & 117 & 142 & \multirow{2}{*}{0} \\
&
&
&
(76.9\%) & (10.4\%) & (12.7\%) & \\

\multirow{2}{*}{Radiologist 12} &
\multirow{2}{*}{119} &
\multirow{2}{*}{3,186} &
2,766 & 320 & 98 & 2 \\
&
&
&
(86.8\%) & (10\%) & (3.1\%) & (0.1\%) \\

\multirow{2}{*}{Radiologist 13} &
\multirow{2}{*}{71} &
\multirow{2}{*}{2,202} &
1,988 & 147 & 67 & \multirow{2}{*}{0} \\
&
&
&
(90.3\%) & (6.7\%) & (3\%) & \\

\multirow{2}{*}{Radiologist 14} &
\multirow{2}{*}{60} &
\multirow{2}{*}{2,117} &
1,685 & 362 & 70 & \multirow{2}{*}{0} \\
&
&
&
(79.6\%) & (17.1\%) & (3.3\%) & \\

\multirow{2}{*}{Radiologist 15} &
\multirow{2}{*}{17} &
\multirow{2}{*}{487} &
369 & 79 & 39 & \multirow{2}{*}{0} \\
&
&
&
(75.8\%) & (16.2\%) & (8\%) & \\

\midrule

\multirow{2}{*}{\textbf{Total}} &
\multirow{2}{*}{\textbf{638}} &
\multirow{2}{*}{\textbf{22,270}} &
\textbf{18,493} &
\textbf{2,729} &
\textbf{1,041} &
\textbf{7} \\
&
&
&
\textbf{(83.04\%)} &
\textbf{(12.25\%)} &
\textbf{(4.67\%)} &
\textbf{($<$0.04\%)} \\

\bottomrule

\end{tabular}%
}

\vspace{0.5em}

\begin{minipage}{\textwidth}
\footnotesize
Percentages may not
total 100\% because of rounding.
\end{minipage}

\end{table}

\newpage
\begin{table}[ht]
\centering

\caption{Report-level QA flags identified by Agent 4 (DeepSeek-R1) across 638 radiology reports, showing the total number of reports reviewed for each radiologist, the number of reports flagged by the QA agent, the severity of flagged errors (critical, major, minor), and the frequency of specific error categories, including section mismatches, gender–anatomy discrepancies, and missing communication of critical findings.}
\label{tab:qa_flags}
\renewcommand{\arraystretch}{1.05}
\setlength{\tabcolsep}{3pt}
\resizebox{\textwidth}{!}{%
\begin{tabular}{@{}ccccccccc@{}}
\toprule

\multirow[c]{3}{*}{\textbf{Radiologist}} &
\multirow[c]{3}{*}{\textbf{Reports, n}} &
\multirow[c]{3}{*}{%
\shortstack[c]{\textbf{Flagged}\\
\textbf{reports,}\\
\textbf{n (\%)}}} &
\multicolumn{3}{c}{\textbf{Flag severity, n (\%)}} &
\multicolumn{3}{c}{\textbf{Error type, n (\%)}} \\

\cmidrule(lr){4-6}
\cmidrule(lr){7-9}

&
&
&
\multirow[c]{2}{*}{\textbf{Critical}} &
\multirow[c]{2}{*}{\textbf{Major}} &
\multirow[c]{2}{*}{\textbf{Minor}} &
\multirow[c]{2}{*}{%
\shortstack[c]{\textbf{Section}\\
\textbf{mismatch}}} &
\multirow[c]{2}{*}{%
\shortstack[c]{\textbf{Gender-anatomy}\\
\textbf{discrepancy}}} &
\multirow[c]{2}{*}{%
\shortstack[c]{\textbf{Missing critical}\\
\textbf{communication}}} \\

&
&
&
&
&
&
&
&
\\

\midrule
\multirow{2}{*}{Radiologist 1} &
\multirow{2}{*}{40} &
11 &
\multirow{2}{*}{0} &
6 &
5 &
10 &
\multirow{2}{*}{0} &
1 \\
&
&
(27.5\%) &
&
(15\%) &
(12.5\%) &
(25\%) &
&
(2.5\%) \\
\multirow{2}{*}{Radiologist 2} &
\multirow{2}{*}{19} &
2 &
\multirow{2}{*}{0} &
2 &
\multirow{2}{*}{0} &
2 &
\multirow{2}{*}{0} &
\multirow{2}{*}{0} \\
&
&
(10.5\%) &
&
(10.5\%) &
&
(10.5\%) &
&
\\
\multirow{2}{*}{Radiologist 3} &
\multirow{2}{*}{42} &
2 &
\multirow{2}{*}{0} &
\multirow{2}{*}{0} &
2 &
2 &
1 &
\multirow{2}{*}{0} \\
&
&
(4.8\%) &
&
&
(4.8\%) &
(4.8\%) &
(2.4\%) &
\\
\multirow{2}{*}{Radiologist 4} &
\multirow{2}{*}{45} &
4 &
1 &
1 &
2 &
3 &
2 &
1 \\
&
&
(8.9\%) &
(2.2\%) &
(2.2\%) &
(4.4\%) &
(6.7\%) &
(4.4\%) &
(2.2\%) \\
\multirow{2}{*}{Radiologist 5} &
\multirow{2}{*}{62} &
6 &
\multirow{2}{*}{0} &
4 &
2 &
6 &
\multirow{2}{*}{0} &
\multirow{2}{*}{0} \\
&
&
(9.7\%) &
&
(6.5\%) &
(3.2\%) &
(9.7\%) &
&
\\
\multirow{2}{*}{Radiologist 6} &
\multirow{2}{*}{27} &
6 &
1 &
2 &
3 &
5 &
\multirow{2}{*}{0} &
1 \\
&
&
(22.2\%) &
(3.7\%) &
(7.4\%) &
(11.1\%) &
(18.5\%) &
&
(3.7\%) \\
\multirow{2}{*}{Radiologist 7} &
\multirow{2}{*}{39} &
4 &
\multirow{2}{*}{0} &
\multirow{2}{*}{0} &
4 &
4 &
\multirow{2}{*}{0} &
\multirow{2}{*}{0} \\
&
&
(10.3\%) &
&
&
(10.3\%) &
(10.3\%) &
&
\\
\multirow{2}{*}{Radiologist 8} &
\multirow{2}{*}{21} &
3 &
1 &
1 &
1 &
2 &
\multirow{2}{*}{0} &
1 \\
&
&
(14.3\%) &
(4.8\%) &
(4.8\%) &
(4.8\%) &
(9.5\%) &
&
(4.8\%) \\
\multirow{2}{*}{Radiologist 9} &
\multirow{2}{*}{29} &
7 &
1 &
3 &
3 &
7 &
\multirow{2}{*}{0} &
1 \\
&
&
(24.1\%) &
(3.4\%) &
(10.3\%) &
(10.3\%) &
(24.1\%) &
&
(3.4\%) \\
\multirow{2}{*}{Radiologist 10} &
\multirow{2}{*}{11} &
1 &
\multirow{2}{*}{0} &
\multirow{2}{*}{0} &
1 &
1 &
\multirow{2}{*}{0} &
\multirow{2}{*}{0} \\
&
&
(9.1\%) &
&
&
(9.1\%) &
(9.1\%) &
&
\\
\multirow{2}{*}{Radiologist 11} &
\multirow{2}{*}{36} &
5 &
\multirow{2}{*}{0} &
2 &
3 &
3 &
\multirow{2}{*}{0} &
\multirow{2}{*}{0} \\
&
&
(13.9\%) &
&
(5.6\%) &
(8.3\%) &
(8.3\%) &
&
\\
\multirow{2}{*}{Radiologist 12} &
\multirow{2}{*}{119} &
22 &
\multirow{2}{*}{0} &
12 &
10 &
22 &
1 &
1 \\
&
&
(18.5\%) &
&
(10.1\%) &
(8.4\%) &
(18.5\%) &
(0.8\%) &
(0.8\%) \\
\multirow{2}{*}{Radiologist 13} &
\multirow{2}{*}{71} &
11 &
1 &
5 &
5 &
9 &
\multirow{2}{*}{0} &
1 \\
&
&
(15.5\%) &
(1.4\%) &
(7\%) &
(7\%) &
(12.7\%) &
&
(1.4\%) \\
\multirow{2}{*}{Radiologist 14} &
\multirow{2}{*}{60} &
3 &
\multirow{2}{*}{0} &
1 &
2 &
3 &
\multirow{2}{*}{0} &
\multirow{2}{*}{0} \\
&
&
(5\%) &
&
(1.7\%) &
(3.3\%) &
(5\%) &
&
\\
\multirow{2}{*}{Radiologist 15} &
\multirow{2}{*}{17} &
3 &
1 &
1 &
1 &
1 &
\multirow{2}{*}{0} &
2 \\
&
&
(17.6\%) &
(5.9\%) &
(5.9\%) &
(5.9\%) &
(5.9\%) &
&
(11.8\%) \\
\midrule
\multirow{2}{*}{\textbf{Total}} &
\multirow{2}{*}{\textbf{638}} &
\textbf{90} &
\textbf{6} &
\textbf{40} &
\textbf{44} &
\textbf{80} &
\textbf{4} &
\textbf{9} \\
&
&
\textbf{(14.1\%)} &
\textbf{(0.9\%)} &
\textbf{(6.3\%)} &
\textbf{(6.9\%)} &
\textbf{(12.5\%)} &
\textbf{(0.6\%)} &
\textbf{(1.4\%)} \\
\bottomrule
\end{tabular}%
}

\vspace{0.5em}
\begin{minipage}{\textwidth}
\footnotesize
Error categories are not mutually exclusive; a single report may contain multiple error types.
\end{minipage}

\end{table}

\newpage
\begin{table}[ht]
\centering
\caption{Average processing time per report across 15 radiologists for radiology report structuring and quality assurance tasks performed by the multi-agent AI pipeline. Columns represent average processing times for each agent, while the rightmost column shows the overall average total processing time per report for each radiologist.}
\label{tab:processing_time}
\renewcommand{\arraystretch}{1.18}
\begin{tabular*}{\textwidth}{@{\extracolsep{\fill}}lccccc@{}}
\toprule
\multirow[b]{3}{*}{\textbf{Radiologist}} &
\multicolumn{4}{c}{\textbf{Average processing time per report (s)}} &
\multirow[b]{3}{*}{%
\shortstack{\textbf{Average total processing}\\
\textbf{time per report (s)}}} \\
\cmidrule(lr){2-5}
&
\multicolumn{3}{c}{\textbf{Structuring}} &
\textbf{QA} &
\\
\cmidrule(lr){2-4}
\cmidrule(lr){5-5}
&
\shortstack{\textbf{Agent 1}\\\textbf{(regex)}} &
\shortstack{\textbf{Agent 2}\\\textbf{(LLaMA 3)}} &
\shortstack{\textbf{Agent 3}\\\textbf{(DeepSeek-R1)}} &
\shortstack{\textbf{Agent 4}\\\textbf{(DeepSeek-R1)}} &
\\
\midrule
Radiologist 1  & 0.538 & 1.467 & 21.078  & 25.094 & 48.177 \\
Radiologist 2  & 0.628 & 2.796 & 45.284  & 24.258 & 72.966 \\
Radiologist 3  & 0.512 & 0.536 & 19.999  & 23.222 & 44.269 \\
Radiologist 4  & 0.758 & 3.310 & 47.074  & 24.061 & 75.203 \\
Radiologist 5  & 0.544 & 1.386 & 6.370   & 24.717 & 33.016 \\
Radiologist 6  & 0.865 & 1.492 & 18.155  & 51.782 & 72.294 \\
Radiologist 7  & 0.401 & 0.911 & 11.945  & 20.511 & 33.769 \\
Radiologist 8  & 0.877 & 2.220 & 28.706  & 30.837 & 62.640 \\
Radiologist 9  & 0.501 & 1.608 & 33.268  & 26.165 & 61.641 \\
Radiologist 10 & 0.849 & 3.447 & 37.402  & 29.000 & 70.697 \\
Radiologist 11 & 0.482 & 0.876 & 92.138  & 21.896 & 115.391 \\
Radiologist 12 & 0.409 & 0.666 & 9.418   & 22.056 & 32.549 \\
Radiologist 13 & 0.479 & 0.536 & 10.674  & 26.682 & 38.371 \\
Radiologist 14 & 0.542 & 1.380 & 16.730  & 22.283 & 40.936 \\
Radiologist 15 & 0.438 & 1.632 & 239.237 & 24.158 & 265.465 \\
\midrule
\textbf{Average}
& \textbf{0.539}
& \textbf{1.310}
& \textbf{28.618}
& \textbf{25.131}
& \textbf{55.598} \\
\bottomrule
\end{tabular*}

\end{table}

\newpage
\begin{table}[htbp]
\centering

\caption{Independent radiologist evaluation of report structuring and
QA performance in 45 reports.}
\label{tab:radiologist_evaluation}

\renewcommand{\arraystretch}{1.08}

\begin{tabular}{@{}L{2.3cm} L{7.0cm} L{4.5cm} L{2.7cm}@{}}

\toprule
\textbf{Task} &
\textbf{Evaluation category} &
\textbf{Answer} &
\textbf{Reports, n (\%)} \\
\midrule


\textbf{Structuring}
&
\multirow[t]{3}{7.0cm}{Newly restructured report is correctly structured}
& Yes
& 31 (69\%) \\

&
&
No
& 2 (4\%) \\

&
&
Evaluator disagreement
& 12 (27\%) \\

\cmidrule(lr){2-4}

&
\multirow[t]{2}{7.0cm}{Clinically important information was omitted from the newly structured report}
& Yes
& 0 \\

&
&
No
& 45 (100\%) \\

\cmidrule(lr){2-4}

&
\multirow[t]{2}{7.0cm}{Fabricated information was introduced in the newly structured report}
& Yes
& 0 \\

&
&
No
& 45 (100\%) \\

\cmidrule(lr){2-4}

&
\multirow[t]{4}{7.0cm}{Newly structured report is better for clinical review than the original report}
& Yes
& 1 (2\%) \\

&
&
About the same
& 23 (51\%) \\

&
&
No
& 3 (7\%) \\

&
&
Evaluator disagreement
& 18 (40\%) \\


\midrule

\textbf{QA}
&
\multirow[t]{4}{7.0cm}{Detection of mismatches between the Findings and Impression sections or within either section}
& Mismatch detected
& 27 (60\%) \\

&
&
No mismatch detected
& 10 (22\%) \\

&
&
False-positive flag
& 5 (11\%) \\

&
&
Evaluator disagreement
& 3 (7\%) \\

\cmidrule(lr){2-4}

&
\multirow[t]{2}{7.0cm}{Detection of gender--anatomy discrepancies}
& False-positive flag
& 1 (2\%) \\

&
&
No discrepancy detected
& 44 (98\%) \\

\cmidrule(lr){2-4}

&
\multirow[t]{4}{7.0cm}{Detection of undocumented communication of critical findings}
& Identified missing communication
& 5 (11\%) \\

&
&
Identified no communication issue
& 38 (84\%) \\

&
&
False-positive flag
& 1 (2\%) \\

&
&
Evaluator disagreement
& 1 (2\%) \\

\cmidrule(lr){2-4}

&
\multirow[t]{3}{7.0cm}{Flag severity}
& Correctly assigned
& 28 (62\%) \\

&
&
Incorrectly assigned
& 2 (4\%) \\

&
&
Evaluator disagreement
& 15 (33\%) \\

\cmidrule(lr){2-4}

&
\multirow[t]{3}{7.0cm}{QA-agent reasoning is correct and clinically reasonable}
& Yes
& 29 (64\%) \\

&
&
Mostly
& 6 (13\%) \\

&
&
Evaluator disagreement
& 10 (22\%) \\

\cmidrule(lr){2-4}

&
\multirow[t]{4}{7.0cm}{Overall QA performance}
& Excellent
& 21 (47\%) \\

&
&
Good
& 2 (4\%) \\

&
&
Fair
& 7 (16\%) \\

&
&
Evaluator disagreement
& 15 (33\%) \\

\bottomrule

\end{tabular}

\vspace{0.5em}

\begin{minipage}{\textwidth}
\footnotesize
Values are numbers of reports, with percentages calculated using the
45-report expert evaluation subset. Evaluator disagreement indicates discordant
assessments between the two reviewers. Percentages may not
total 100\% because of rounding.
\end{minipage}

\end{table}


\end{document}